\documentclass[letterpaper]{article} 
\usepackage{aaai23}  
\usepackage{times}  
\usepackage{helvet}  
\usepackage{courier}  
\usepackage[hyphens]{url}  
\usepackage{graphicx} 
\usepackage{natbib}  
\usepackage{caption} 
\usepackage{booktabs}
\usepackage{color}
\usepackage{xcolor}
\usepackage{mathrsfs}
\usepackage{multirow}
\usepackage{bbding}  
\usepackage{amsthm,amsmath,amssymb}  
\usepackage{makecell} 

\usepackage[ruled,linesnumbered]{algorithm2e}

\usepackage{newfloat}
\usepackage{listings}
\DeclareCaptionStyle{ruled}{labelfont=normalfont,labelsep=colon,strut=off} 
\title{RankDNN: Learning to Rank for Few-shot Learning}

\author{
	Qianyu Guo\textsuperscript{\rm 1,2}, 
 	Haotong Gong\textsuperscript{\rm 1}, 
        Xujun Wei\textsuperscript{\rm 1,3}, 
        Yanwei Fu\textsuperscript{\rm 2}, 
 	Weifeng Ge\textsuperscript{\rm 1,2\footnote{Corresponding author}}
        Yizhou Yu\textsuperscript{\rm 4}, 
	Wenqiang Zhang\textsuperscript{\rm 2,3}
	\\
}

\affiliations{
 \textsuperscript{\rm 1}Nebula AI Group, School of Computer Science, Fudan University,Shanghai,China\\
	\textsuperscript{\rm 2}Shanghai Key Laboratory of Intelligent Information Processing,Shanghai,China \\ 
	\textsuperscript{\rm 3}Academy for Engineering \& Technology, Fudan University,Shanghai,China\\
   	\textsuperscript{\rm 4}Department of Computer Science, The University of Hong Kong,Hong Kong,China\\
	\
wfge@fudan.edu.cn

}

\usepackage{bibentry}

\begin{document}

\maketitle

\begin{abstract}
This paper introduces a new few-shot learning pipeline that casts relevance ranking for image retrieval as binary ranking relation classification. In comparison to image classification, ranking relation classification is sample efficient and domain agnostic. Besides, it provides a new perspective on few-shot learning and is complementary to state-of-the-art methods. The core component of our deep neural network is a simple MLP, which takes as input an image triplet encoded as the difference between two vector-Kronecker products, and outputs a binary relevance ranking order. The proposed RankMLP can be built on top of any state-of-the-art feature extractors, and our entire deep neural network is called the ranking deep neural network, or RankDNN. Meanwhile, RankDNN can be flexibly fused with other post-processing methods. During the meta test, RankDNN ranks support images according to their similarity with the query samples, and each query sample is assigned the class label of its nearest neighbor. Experiments demonstrate that RankDNN can effectively improve the performance of its baselines based on a variety of backbones and it outperforms previous state-of-the-art algorithms  on multiple few-shot learning benchmarks, including miniImageNet, tieredImageNet, Caltech-UCSD Birds, and CIFAR-FS. Furthermore, experiments on the cross-domain challenge demonstrate the superior transferability of RankDNN.The code is available at: https://github.com/guoqianyu-alberta/RankDNN.
\end{abstract}

\section{Introduction}
Contrary to the normal practice of using a large amount of labeled data~\cite{krizhevsky2012imagenet,he2016deep}, few-shot learning~\cite{lu2020learning,afrasiyabi2021mixture} refers to learning new concepts from a very small amount of data by leveraging the learning ``skills" gained from many similar learning tasks. State-of-the-art methods~\cite{ma2021partner,hong2021reinforced,tang2021mutual,rizve2021exploring} attempt to learn meta knowledge that can be transferred to new tasks. Although much progress has been made, recently proposed methods still bear the risk of unsatisfactory generalization performance on new tasks. The reason is two-fold. First, the meta knowledge they learn may have limited transferability and may not be well applicable to certain new tasks. Second, there is simply too little data available for new tasks and overfitting is very hard to avoid. Modern deep neural networks are so flexible and overparameterized that they can even perfectly fit image data with random labels, as stated in~\cite{zhang2021understanding,arpit2017closer}. Overfitting creates a large gap between training and testing performance. 

\begin{figure}[t]
	\centering
	\includegraphics[width=1.0\linewidth]{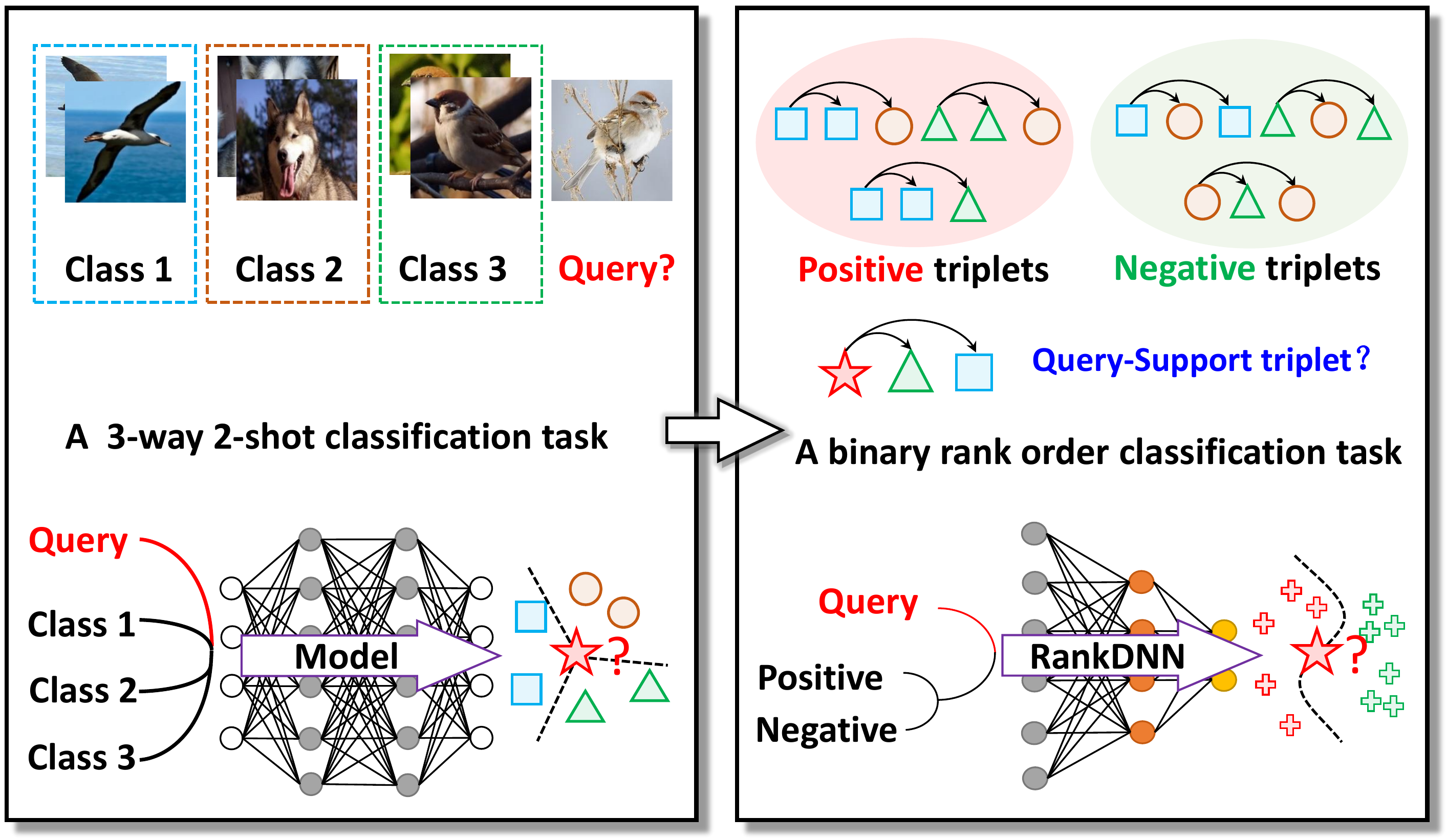}
	\caption{ {Ranking deep neural network (RankDNN) establishes a learning framework that can cast a $N$-way-$K$-shot learning task into a binary relevance ranking problem, which is sample efficient and domain agnostic. } }
	\label{Fig:motivation}
\end{figure}

\begin{figure*}[t]
	\centering
	\includegraphics[width=0.9\linewidth]{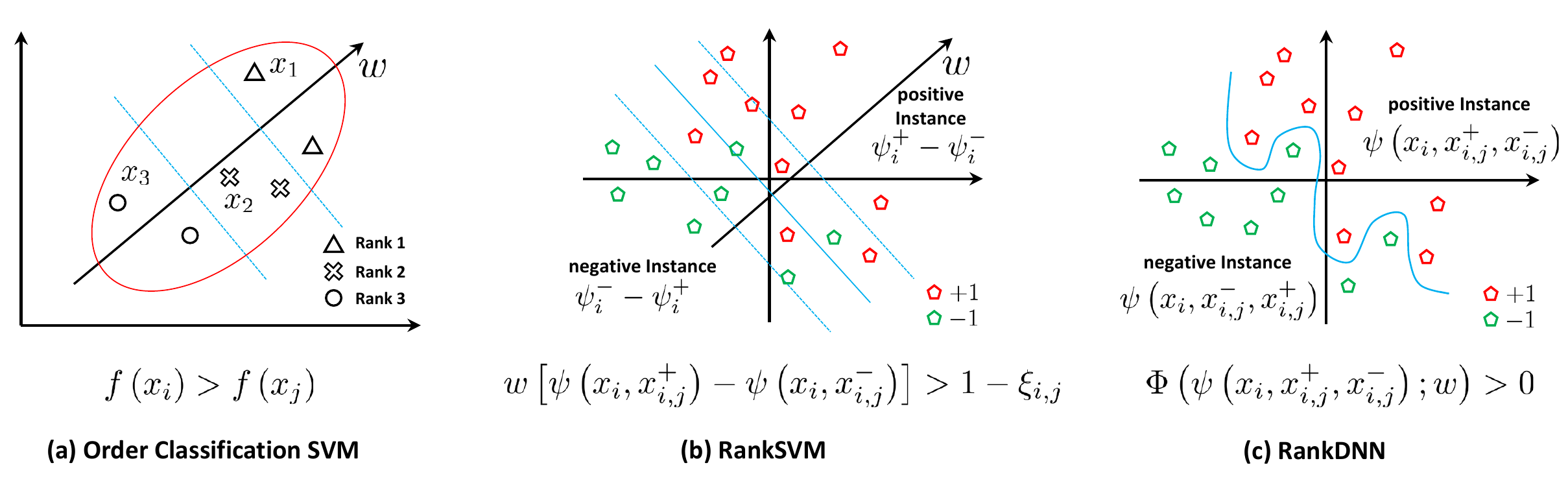}
	\caption{Comparison of ranking learning frameworks. (a) Pointwise ranking algorithms, such as McRank~\cite{li2007mcrank,shashua2003ranking} and OC SVM, cast the ranking learning problem as a regression or classification task on single objects. (b) Pairwise ranking algorithms, such as RankSVM~\cite{joachims2002optimizing,prosser2010person}, models binary relevance ranking within object pairs. (c) The proposed RankDNN generalizes binary relevance ranking in RankSVM into binary triplet classification with deep neural networks.}
	\label{Fig:rank_problem}
\end{figure*}

Inspired by relevance ranking in information retrieval~\cite{shashua2003ranking,joachims2002optimizing}, we consider binary relevance ranking for few-shot learning, as shown in  Fig~\ref{Fig:motivation}. Given a query image and two support images in a given order, binary relevance ranking ranks the relevance of these support images concerning the query image. There are only two possible outcomes of this ranking problem. That is, the first support image is more relevant than the second one or vice versa. Relevance ranking among images focuses on the similarity of images, especially the similarity of foreground objects, but not the specific categories of images or foreground objects. Thus a relevance ranking model should be able to grasp the meta knowledge about the assessment of image or object similarity regardless of their specific contents. Such a capability endows the learned model a strong transferability to a wide range of new tasks where specific image contents may vary. Meanwhile, as we need three images for each binary relevance ranking instance, the number of training samples for ranking is greatly expanded by constructing triplets, which significantly alleviates training data shortage in few-shot learning.

We develop a ranking learning framework for few-shot image classification, which is formulated as a query-support relevance ranking problem. We generalize the SVM based ranking learning idea in OC SVM~\cite{shashua2003ranking} and RankSVM~\cite{joachims2002optimizing} to deep neural networks, and call the proposed framework RankDNN. 
Meanwhile, we provide a new perspective on image classification by converting a multiclass classification task into a binary classification task.

To implement binary relevance ranking of two support images to a query image, there are two key issues: The first issue is how to simultaneously encode the semantic features of an image triplet as well as the roles and order of the three images in the triplet; the second issue is which machine learning algorithm should be used to produce the binary ranking result. Encoding an image triplet while preserving all necessary information is very challenging. We use the query image in the triplet to form two query-support pairs with the two support images, and encode each query-support pair using the outer product, also called vector-Kronecker product, of their feature vectors. The entire image triplet is finally represented using the difference between these two vector outer products.
Our intuition behind this encoding scheme is that the vector-Kronecker product of two  feature vectors can not only preserve all information in the original feature vectors but also model the correlation between any two dimensions of these features. In addition, the sign of the final difference encodes the order of the two support images in the triplet. 

In summary, this paper has the following contributions:
{\flushleft $\bullet$} We introduce a new ranking learning framework for few-shot learning called ranking deep neural network (or RankDNN), which decomposes a few-shot image classification task into multiple binary relevance ranking problems. By constructing image triplets for such binary ranking problems, our framework generates a large amount of training data across different image classes.

{\flushleft $\bullet$} We propose a novel image triplet encoding scheme based on vector-Kronecker products. It preserves all necessary information in an image triplet and can also model the correlation between any two feature dimensions respectively of the query image and one of the support images. 

{\flushleft $\bullet$} Experiments demonstrate that our RankDNN outperforms its baselines in many different scenarios
and achieves state-of-the-art performance on multiple few-shot learning benchmarks
, including {\em mini}ImageNet, ~\cite{vinyals2016matching}, {\em tiered}ImageNet, ~\cite{ren2018meta}, Caltech-UCSD Birds~\cite{wah2011caltech}, and CIFAR-FS. \cite{bertinetto2018meta}. 
Besides, the fusion experiments, cross-domain challenges and different backbones experiments prove the generalization, flexibility and robustness of RankDNN.



\begin{figure*}[t]
	\centering
	\includegraphics[width=0.98\linewidth]{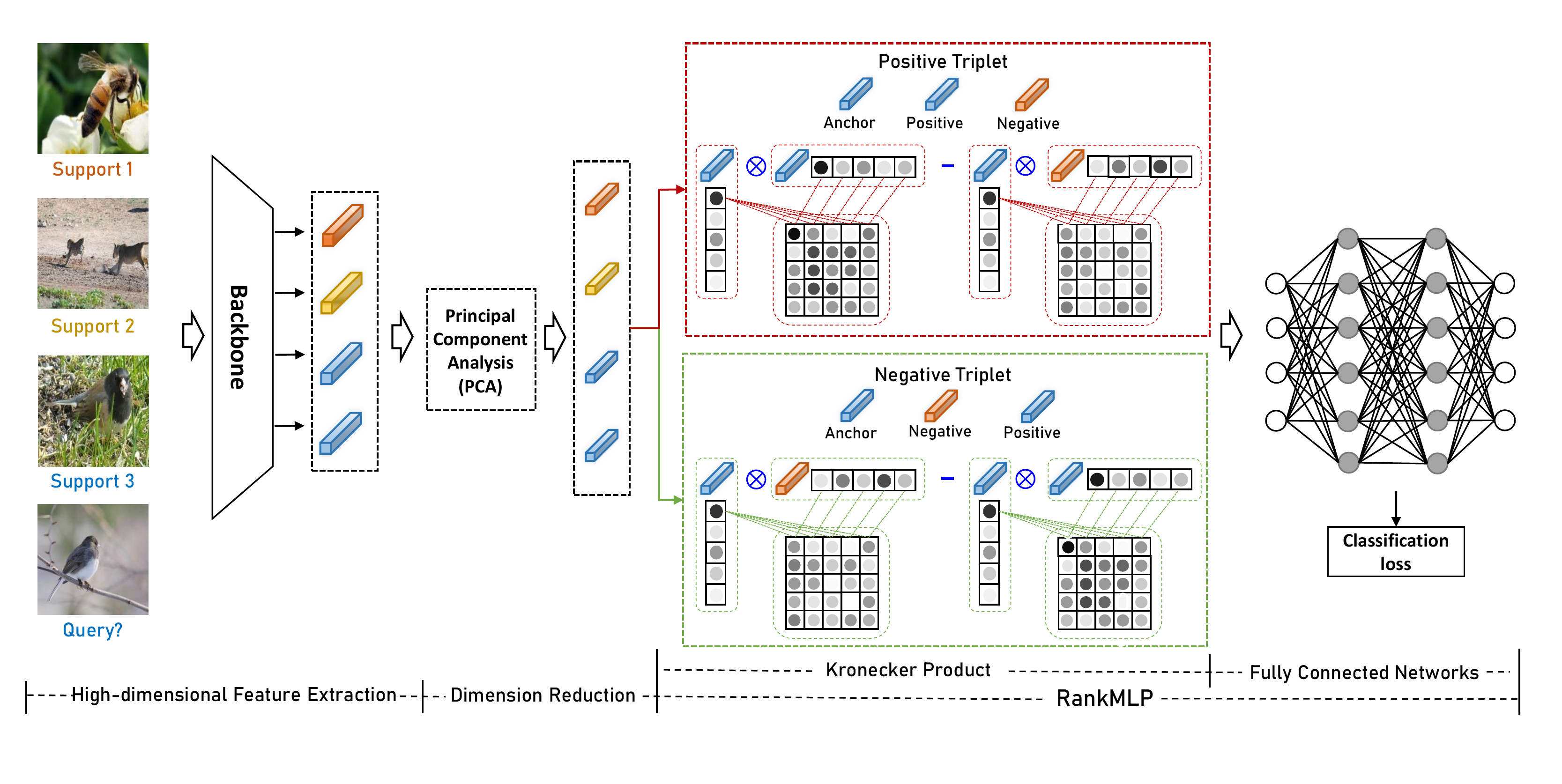}
	\caption{{The ranking deep neural network is composed of a high-dimensional feature extractor, a dimension reduction part and a ranking multilayer preceptron. The parameters feature extractor are frozen all the time and the RankMLP are trained consecutively in different stages.}}
	\label{Fig:pipeline}
\end{figure*}

\section{Related Work}
{\noindent\textbf{Few-shot Learning.} 
Few-shot learning aims to recognize samples from unseen classes with limited training data, which is expect to reduce labeling cost, achieve a low-cost and quick model deployment. 
Various few-shot learning algorithms have been proposed, such as Matching Net~\cite{vinyals2016matching}, Relation Net~\cite{sung2018learning}, MAML~\cite{finn2017model}, and TAML~\cite{jamal2019task}. However, according to \cite{chen2019closerfewshot,gidaris2018dynamic,qiao2018few}, since there are two few annotated samples, state-of-the-art algorithms~\cite{mangla2020charting,rizve2021exploring} focus on solving the over-fitting problem by self-supervised learning. There are also many other methods that try to learn meta knowledge with the strong generalization ability~\cite{ hong2021reinforced,tang2021mutual}.In this paper, we provide a new perspective to solve the few-shot learning problem by converting it into a relevance ranking task. Then our learning task is built upon image triplets and becomes domain agnostic. Since there are a large amount of triplet combinations even with few samples, the overfitting problem can be alleviated greatly.

\noindent\textbf{Learning to Rank.}
Applying existing and effective machine learning methods to rank is called learning to rank, such as RankSVM\cite{joachims2002optimizing} and RankNet\cite{burges2005learning}. Learning to rank is not only widely used in major tasks of natural language processing\cite{masuda2003ranking,briakou2020detecting,zhang2020translation}, but also has attracted the attention of metric learning in recent years\cite{liu2021noise,cakir2019deep,wang2019ranked}. \cite{wang2019ranked} proposed ranked-based list loss with structured information of multiple samples to improve the training efficiency of metric learning.
Although DeepRank~\cite{pang2017deeprank} utilizes deep neural networks to conduct similarity ranking, its learning objective is a pairwise contrastive loss which is widely used in metric learning. While in RankDNN, we convert the rank problem into a binary classification problem and solve it with gradient-based methods. 
Meanwhile, compared with that in~\cite{prosser2010person}, we propose a much more efficient triplet feature the construction method for few-shot learning and solve it with deep neural network flexibly.

\section{A Ranking Learning Framework for Few-shot Image Classification}

The general regime of an {\em N}-way {\em K}-shot classification task $\mathcal{T}$:
with $N$ previously unseen image categories, each of which contains $K$ samples, the task aims to build a classifier to classify a set of query images into these $N$ categories.
The dataset for task $\mathcal{T}$ is $\mathcal{D}_{\mathcal{T}} = \mathcal{S} \cup \mathcal{Q}$ where $\mathcal{S} = \left \{ \left ( \boldsymbol{x}_i, y_i \right ) \right \} _{i=1}^{N \times K}$ is the support set, $\mathcal{Q} = \left \{ \left ( \boldsymbol{x}_i, y_i \right ) \right \} _{i= N \times K + 1}^{N \times K + T}$ is the query set, $T$ is the number of query samples. $\boldsymbol{x}_i$ and $y_i (\in \left \{ C_1,...,C_N \right \} = \mathcal{C}_{\mathcal{T}}\subset \mathcal{C}$) are respectively the $i$-the image sample and its label. State-of-the-art methods often formalize few-shot image classification as a metric learning~\cite{zhang2020deepemd,mangla2020charting} or multiclass classification task~\cite{rizve2021exploring,chen2021meta,yang2021free}. However, in this paper, we cast few-shot classification as an image retrieval task. Given a query image $\boldsymbol{x}_q$ and a support image collection $\mathcal{S}$, a retrieval system should return a complete ranked list $r^*(\boldsymbol{x}_q)$, that sorts the support images in $\mathcal{S}$ according to their relevance to the query as follows.
\begin{equation}
	\begin{aligned}
		r^*(\boldsymbol{x}_q): \boldsymbol{x}_{q_1} \succ \boldsymbol{x}_{q_2} \succ \cdots \succ \boldsymbol{x}_{q_{N\times K}},
	\end{aligned}
	\label{eq:cls2rank}
\end{equation}
where $\boldsymbol{x}_{q_k} \in \mathcal{S}$, and  $\boldsymbol{x}_{q_i} \succ \boldsymbol{x}_{q_j}$ means $\boldsymbol{x}_{q_i}$ is more relevant to $\boldsymbol{x}_q$ than $\boldsymbol{x}_{q_j}$. The formula in (\ref{eq:cls2rank}) indicates how an image retrieval system works~\cite{ge2018deep,wang2020cross,levi2021rethinking}. Most metric learning based methods~\cite{zhang2020deepemd,snell2017prototypical} for few-shot learning aim to find a unified embedding space where every query sample is assigned the class label of its nearest support image.

Inspired by the approach in~\cite{joachims2002optimizing,prosser2010person}, we decompose the complete relevance ranking relation among multiple variables into several binary relevance ranking relations over $\mathcal{S} \times \mathcal{S}$. Thus, we have
\begin{equation}
	\begin{aligned}
		\forall \left ( \boldsymbol{x}_{q_i}, \boldsymbol{x}_{q_j}  \right ) \in r^*\left ( \boldsymbol{x}_q \right ) (i<j): \boldsymbol{x}_{q_i} \succ \boldsymbol{x}_{q_j}, \boldsymbol{x}_{q_j} \prec \boldsymbol{x}_{q_i}.
	\end{aligned}
	\label{eq:rankpair}
\end{equation}
There are $\mathcal{O}\left ( \left | \mathcal{S} \right |^2 \right )$ such relevance ranking pairs. The following 5-way-1-shot task is given as an example:
\begin{equation}
	\begin{aligned}
		{\color{red}\boldsymbol{x}_{q_1} \succ \boldsymbol{x}_{q_2}}, {\color{red}\boldsymbol{x}_{q_1} \succ \boldsymbol{x}_{q_3}}, {\color{red}\boldsymbol{x}_{q_1} \succ \boldsymbol{x}_{q_4}}, {\color{red}\boldsymbol{x}_{q_1} \succ \boldsymbol{x}_{q_5}}; \\
		{\color{blue}\boldsymbol{x}_{q_2} \prec \boldsymbol{x}_{q_1}}, {\color{red}\boldsymbol{x}_{q_2} \succ \boldsymbol{x}_{q_3}}, {\color{red}\boldsymbol{x}_{q_2} \succ \boldsymbol{x}_{q_4}}, {\color{red}\boldsymbol{x}_{q_2} \succ \boldsymbol{x}_{q_5}}; \\
		{\color{blue}\boldsymbol{x}_{q_3} \prec \boldsymbol{x}_{q_1}}, {\color{blue}\boldsymbol{x}_{q_3} \prec \boldsymbol{x}_{q_2}}, {\color{red}\boldsymbol{x}_{q_3} \succ \boldsymbol{x}_{q_4}}, {\color{red}\boldsymbol{x}_{q_2} \succ \boldsymbol{x}_{q_5}}; \\
		{\color{blue}\boldsymbol{x}_{q_4} \prec \boldsymbol{x}_{q_1}}, {\color{blue}\boldsymbol{x}_{q_4} \prec \boldsymbol{x}_{q_2}}, {\color{blue}\boldsymbol{x}_{q_4} \prec \boldsymbol{x}_{q_3}}, {\color{red}\boldsymbol{x}_{q_4} \succ \boldsymbol{x}_{q_5}}; \\
		{\color{blue}\boldsymbol{x}_{q_5} \prec \boldsymbol{x}_{q_1}}, {\color{blue}\boldsymbol{x}_{q_5} \prec \boldsymbol{x}_{q_2}}, {\color{blue}\boldsymbol{x}_{q_5} \prec \boldsymbol{x}_{q_3}}, {\color{blue}\boldsymbol{x}_{q_5} \prec \boldsymbol{x}_{q_4}}. \\
	\end{aligned}
	\label{eq:rankpair_example}
\end{equation}
The query image $\boldsymbol{x}_q$ is then assigned the class label of the support sample ranked first. By exploiting such binary relations, we can obtain an extensive number of training samples in few-shot learning, avoid the necessity to obtain a complete ranked list, and tolerate minor inconsistencies in binary ranking results.


According to Eq. (\ref{eq:rankpair}), we need to formulate the binary relevance ranking relation between any two support images $(\boldsymbol{x}_{i},\boldsymbol{x}_{j})$ for a query sample $\boldsymbol{x}_{q}$. There are two key elements in describing such a relation: First, an informative encoding scheme $\psi\left ( \cdot ;\theta \right )$ ($\theta$ denotes the parameters of $\psi$) to encode the semantic features as well as the detailed component-wise correlations between a query image and a support image; second, a scoring function $\Phi$ to evaluate the overall similarity or relevance between a query image and a support image. Then a binary relevance ranking relation can be concretely formulated as follows:
\begin{equation}
	\begin{aligned}
		\boldsymbol{x}_{i} \succ _{r^*(q)} \boldsymbol{x}_{j} \iff \Phi\left ( \psi\left ( x_q,x_i \right) ;w \right )  >   \Phi\left ( \psi\left ( x_q,x_j \right) ;w \right ),
	\end{aligned}
	\label{eq:rankfunction}
\end{equation}
where $w$ denotes the learnable parameters of the scoring function. RankSVM~\cite{joachims2002optimizing} adopts the class of linear scoring functions, and the binary ranking learning problem becomes finding the weight vector that fulfills the most inequalities in Eq.~\ref{eq:rankpair}, which can be formulated as follows.
\begin{equation*}
	\begin{aligned}
		\min\limits_{w,\vec{\xi}} \frac{1}{2} \left \|  w \right \|^2 + C \sum\limits_{i,j,k} \xi_{i,j,k}
	\end{aligned}
	\label{eq:ranksvm1}
\end{equation*}
\begin{equation}
	\begin{aligned}
		\textup{s.t.}~~  &\forall \boldsymbol{x}_{i} \succ_{r^*\left ( q \right )}  \boldsymbol{x}_{j}: w \left [ \psi\left ( x_q,x_i \right) - \psi\left ( x_q,x_j \right)  \right ]   >  1 - \xi_{i,j,q}   \\
		& \forall i  \forall j \forall q: \xi_{i,j,q} > 0,
	\end{aligned}
	\label{eq:ranksvm2}
\end{equation}
where $C$ is a constant, $\xi_{i,j,q}$ is a slack variable. RankSVM~\cite{joachims2002optimizing} can be extended to use non-linear scoring functions via kernel methods.


In  Fig~\ref{Fig:rank_problem}, the few-shot classification problem can now be cast as a binary classification task and solved by RankSVM. To step further, we wish to obtain a more general formulation about the ranking learning problem. Given a query sample $x_q$, there are two kinds of ranking relations: if $\Phi\left ( \psi\left ( x_q,x_i \right) ;w \right )  >   \Phi\left ( \psi\left ( x_q,x_j \right) ;w \right )$, $\left\langle x_q,x_{i},x_{j} \right\rangle$ is identified as a positive triplet sample; otherwise, $\Phi\left ( \psi\left ( x_q,x_i \right) ;w \right )  \leq   \Phi\left ( \psi\left ( x_q,x_j \right) ;w \right )$, and $\left \langle x_q,x_{i},x_{j} \right \rangle$ is a negative triplet sample. Then we can define a unified triplet encoding scheme, which not only encodes the feature vectors of the query and support samples in the triplet, but also their roles and order in the triplet as well as their interactions. In a training triplet, one of the support samples should come from the same class as the query sample while the other support sample should come from a different class. We can automatically generate the ground-truth binary class label of a triplet as follows,
\begin{equation}
	\begin{aligned}
		y\left ( \left \langle x_q,x_{i},x_{j} \right \rangle \right )= \{ \begin{matrix}
			+1, \textup{if}~~ {x}_{i} \succ_{r^*\left ( q \right )}  {x}_{j} ;\\
			-1, \textup{if}~~ {x}_{i} \prec_{r^*\left ( q \right )}  {x}_{j}.
		\end{matrix}
	\end{aligned}
	\label{eq:tripletlabel}
\end{equation}
In this paper, we generalize the linear/nonlinear scoring function $\Phi$ in RankSVM to an MLP based binary classifier, called RankMLP. RankMLP is trained using the following overall loss function,
\begin{equation}
	\begin{aligned}
		\mathcal{L}_{rank} = \frac{1}{N_t}\sum \limits_{q,i,j} \mathcal{L}_{bce}\left ( \Phi\left ( \psi \left (  \left \langle x_q,x_{i},x_{j} \right \rangle; \theta \right ); w \right ), y_{q,i,j}  \right ),
	\end{aligned}
	\label{eq:rankdnn_objective}
\end{equation}
where $N_t$ is the number of triplets in a mini-batch, $\mathcal{L}_{bce}$ is the binary cross-entropy loss, and $y_{q,i,j}$ is the ground-truth binary class label of a triplet.

\noindent\textbf{Discussion.} According to Eqs. (\ref{eq:tripletlabel}) and (\ref{eq:rankdnn_objective}), we solve the ranking learning problems with deep neural networks, as in information retrieval~\cite{palangi2016deep,pang2017deeprank} and metric learning~\cite{cakir2019deep,liu2021noise}. The main difference is that RankMLP focuses on binary ranking relation classification but not the distance constraints in metric learning. Given an $N$-way-$K$-shot task in the training set of a few-shot learning problem, we can construct $(N-1)K(K-1)$ training triplets for every query sample, which significantly alleviates the data shortage.


\section{The Ranking Deep Neural Network}
\subsection{Overview}
The ranking deep neural network (RankDNN) takes in an image triplet and outputs whether this triplet is a positive sample or a negative sample. Although some existing deep metric learning methods~\cite{cakir2019deep,liu2021noise} already use the ranking loss, a key difference here is that inter-class variations are much more challenging in few-shot learning than those in image retrieval. In our experiments, directly training a deep neural network from end to end to classify triplets would lead to gradient explosion. In Fig~\ref{Fig:pipeline}, the RankDNN pipeline contains two parts: a frozen high-dimensional feature extractor $f \left ( \cdot, \boldsymbol{\theta }_e \right )$ and a ranking deep neural network $\Phi \left ( \cdot, \boldsymbol{\theta } \right )$.

Algorithm~\ref{alg:meta_training} outlines the training phase of RankDNN for few-shot learning. Given an image triplet $\left \langle x_q,x_{i},x_{j} \right \rangle$, RankDNN uses state-of-the-art feature extractors, such as S2M2~\cite{mangla2020charting} and IE~\cite{rizve2021exploring}, to extract discriminative features $f\left ( x_q; \boldsymbol{\theta }_e \right ), f\left ( x_i; \boldsymbol{\theta }_e \right ),f\left ( x_j; \boldsymbol{\theta }_e \right ) \in \mathbb{R}^{640}$. For simplicity, we write $f\left ( x_q; \boldsymbol{\theta }_e \right )$ as $f_q$. To generate subsequent features for triplet classification, we use PCA~\cite{yang2004two} to reduce the image feature dimension from 640 to 80.
During the training stage, we freeze both the feature extractor and the PCA transform, and train RankMLP with the loss function $\mathcal{L}_{rank}$ defined in Eq.~(\ref{eq:rankdnn_objective}).

\begin{algorithm}[t]
\caption{Meta-Training of RankDNN for Few-shot Learning}
\label{alg:meta_training}
\LinesNumbered
\KwIn{Training data $\mathcal{D} = \left \{  \left (  \boldsymbol{x}_i, y_i \right ) \right \}_{k=1}^{N}$. Network $f \left ( \cdot, \boldsymbol{\theta }_e \right )$ initialized with a state-of-the-art pretrained model. The ranking neural network $\Phi \left ( \cdot, \boldsymbol{\theta } \right )$ initialized with randomly noises.}
\KwOut{The learnable parameters $\boldsymbol{\theta }$ of the neural networks $\Phi \left ( \cdot, \boldsymbol{\theta } \right )$.}
{\textcolor[rgb]{0.5,0.5,0.5}{// conduct PCA transform for the dimension reduction}} \\
Extract features of all training images and learn a principal component analysis transform (PCA) $T$;\\
{\textcolor[rgb]{0.5,0.5,0.5}{// train the ranking neural network $\Phi$ and freeze the feature extractor $f_e$ all the time}} \\
\While{not converge}{
$t \leftarrow t+1$;\\
Sample anchors randomly and their neighborhoods according to the method in HTL~\cite{ge2018deep};\\
Extract features with $f_e \left ( \cdot, \boldsymbol{\theta }_e \right )$, and freeze it;\\
Reduce the feature dimension with PCA;\\
Construct the triplet ranking description with Eq.~\ref{eq:rank_description} and pass it through the RankMLP $\Phi \left ( \cdot, \boldsymbol{\theta } \right )$;\\
Compute the binary cross entropy loss in a mini-batch  $\mathcal{L}_{rank}$ with Eq.~\ref{eq:rankdnn_objective};\\	Backpropagate the gradients produced at the loss layer and update the learnable parameters $\boldsymbol{\theta }$.
	}
\end{algorithm}

\subsection{Triplet Encoding Scheme}
Encoding triplets of extracted features plays a precursory role for RankMLP. The Kronecker product is a matrix algebra operation that produces structured descriptions for modeling complex systems. If $A \in \mathbb{R}^{n_a \times m_a}$ and $B \in \mathbb{R}^{n_a \times m_a}$, their Kronecker product is denoted by $A \otimes B$ and defined as follows:
\begin{equation*}
	A \otimes B \equiv {\begin{bmatrix} {a_{11}\bf{B}} &\cdots & {a_{1n}\bf{B}} \\ \vdots & \ddots & \vdots \\ {a_{n1}\bf{B}} &\cdots & {a_{nn}\bf{B}} \end{bmatrix}}.
	\label{eq:Kronecker_product}
\end{equation*}
When both matrices degenerate to vectors $ x \in {\mathbb{R}^{d \times {1}}}$ and $ y \in {\mathbb{R}^{1 \ times {d}}}$, their Kronecker product is actually the same as their outer product; and is called vector-Kronecker product in this paper. It is formulated in the following equation:
\begin{equation}
	\bf {x} \otimes \bf {y} \equiv {\begin{bmatrix} {x_{1}y_{1}} &\cdots & {x_{1}y_{n}} \\ \vdots & \ddots & \vdots \\ {x_{n}y_{1}} &\cdots & {x_{n}y_{n}} \end{bmatrix}}.
	\label{eq:vector_Kronecker_product}
\end{equation}
In the literature~\cite{de1994strong,langville2004kronecker,broxson2006kronecker}, the Kronecker product of graphs is one of the usual names of the categorical product of graphs, also called the tensor product. This product of graphs was studied by various authors~\cite{azevedo2020performance,weichsel1962kronecker,mamut2008vertex}, who proved that Kronecker product can encode the connectivity and relationship between graphs.

Given a triplet of feature descriptors $\left \langle f_q, f_{i}, f_{j} \right \rangle$, state-of-the-art methods~\cite{chen2021meta,rizve2021exploring} calculate cosine similarities, $f_q f_i / (\left \| f_q  \right \| \left \| f_i  \right \|)$ and $f_q f_j / (\left \| f_q  \right \| \left \| f_j  \right \|)$, and further rank the similarities. In fact, this cosine similarity 
focuses on the correlation between corresponding entries in the two feature vectors, and may not be able to capture all types of correlations between two vectors. Therefore, we design the following encoding scheme for feature triplets:
\begin{equation}
	\psi \left (  \left \langle x_q,x_{i},x_{j} \right \rangle; \theta \right ) = {f_q} \otimes {f_i} - {f_q} \otimes {f_j}.
	\label{eq:rank_description}
\end{equation}
Note that popular backbones in few-shot learning include ResNet12~\cite{rizve2021exploring},ResNe-18~\cite{ye2020few} and WRN-28-10~\cite{gidaris2019boosting,zagoruyko2016wide}, both of which output 640-dimensional features. Then the result $\psi \left (  \left \langle x_q,x_{i},x_{j} \right \rangle; \theta \right ) \in \mathbb{R}^{d \times d}$ in Eq.~(\ref{eq:rank_description}) would have 409600 dimensions, which is too high. Therefore, we reduce the dimension of each extracted feature vector to 80 with PCA~\cite{yang2004two}.
\begin{table*}[thpb]
\caption{Comparison of 5-way few-shot accuracies on {\em mini}ImageNet and {\em tiered}ImageNet with ResNet backbones.}
	\centering
	\label{tab:backbone mini/tiered}
	\scalebox{1.0}{
\begin{tabular}{lccccc}
\bottomrule
\multirow{1}{*}{\bf {Method}} &\bf {Backbone} &\multicolumn{2}{c}{\bf {{\em mini}ImagNet}} &\multicolumn{2}{c}{\bf {{\em tiered}Imagenet}} \\
  \cmidrule(r){3-4}  \cmidrule(r){5-6} 
& &1-shot &5-shot &1-shot &5-shot\\
\hline
DMF$_{\;\rm{\,CVPR21}}$
&ResNet-12 &$67.76_{\pm 0.46}$ &$82.71_{\pm 0.31}$ &$71.89_{\pm 0.52}$ &$85.96_{\pm 0.35}$\\
IE$_{\;\rm{\,CVPR21}}$ 
&ResNet-12 &$67.28_{\pm 0.80}$ &$84.78_{\pm 0.52}$ &$72.21_{\pm 0.90}$ &$87.08_{\pm 0.58}$\\
DMN4$_{\;\rm{\,AAAI22}}$ 
&ResNet-12&$66.58$ &$83.52$  &$72.10$ &$85.72$\\
HGNN$_{\;\rm{\,AAAI22}}$
&ResNet-12&$67.02_{\pm 0.20}$ &$83.00_{\pm 0.13}$ &$72.05_{\pm 0.23}$ &$86.49_{\pm 0.15}$\\
APP2S$_{\;\rm{\,AAAI22}}$
&ResNet-12&$66.25_{\pm 0.20}$ &$83.42_{\pm 0.15}$ &$72.00_{\pm 0.22}$ &$86.23_{\pm 0.15}$\\
UNICORN$_{\;\rm{\,ICLR22}}$
&ResNet-12&$65.17_{\pm 0.20}$ &$84.30_{\pm 0.13}$ &$69.24_{\pm 0.20}$ &$86.06_{\pm 0.16}$\\
TAS$_{\;\rm{\,ICLR22}}$
&ResNet-12&$65.68_{\pm 0.45}$ &$83.92_{\pm 0.55
}$ &$72.81_{\pm 0.48}$ &$86.06_{\pm 0.16}$\\
{\bf RankDNN(ours)} &ResNet-12 &${\bf 68.72_{\pm 0.15}}$ &${\bf 85.66_{\pm 0.27}}$ &${\bf 73.87_{\pm 0.40}}$ &${\bf 87.92_{\pm 0.15}}$\\
\hline

$\bigtriangleup-$encode$_{\;\rm{\,NIPS18}}$
&ResNet-18 &$59.90$ &$69.70$ &| &|\\
SS$_{\;\rm{\,ECCV20}}$
&ResNet-18 &| &$76.60$ &| &$78.90$\\
FEAT$_{\;\rm{\,CVPR20}}$
&ResNet-18 &55.15&66.78 &62.40 &77.81\\
Hyperbolic$_{\;\rm{\,CVPR20}}$
&ResNet-18 &$57.05_{\pm 0.20}$&$76.84_{\pm 0.14}$ &$66.20_{\pm 0.28}$ &$76.50_{\pm 0.40}$\\
{\bf RankDNN(ours)} &ResNet-18 &${\bf 62.52_{\pm 0.25}}$ &${\bf 77.33_{\pm 0.45}}$ &${\bf 66.97_{\pm 0.18}}$ &${\bf 81.07_{\pm 0.23}}$\\
\hline
S2M2$_{\;\rm{\,WACV20}}$
&WRN &$64.93_{\pm 0.18}$&$83.18_{\pm 0.11}$ &$73.71_{\pm 0.22}$ &$88.59_{\pm 0.14}$\\
FEAT$_{\;\rm{\,CVPR20}}$
&WRN &65.10  &81.11 &70.41 &84.38\\
PSST$_{\;\rm{\,CVPR21}}$
&WRN &$64.16_{\pm 0.44}$ &$80.64_{\pm 0.32}$ &| &|\\

{\bf RankDNN(ours)} &WRN &${\bf  66.67_{\pm 0.15}}$ &${\bf 84.79_{\pm 0.11}}$ &${\bf 74.00_{\pm 0.15}}$ &${\bf 88.80_{\pm 0.25}}$\\
\bottomrule

\end{tabular}}
\label{tab:mini_and_tiered}
\end{table*}


\section{Experiments}
\label{sec:Experiments}
\subsection{Experimental Details}

\subsection{Datasets and Performance Metrics}
We use four popular benchmark datasets in our experiments: {\em mini}ImageNet\cite{vinyals2016matching}, {\em tiered}ImageNet~\cite{ren2018meta}, Caltech-UCSD Birds-200-2011 (CUB)\cite{wah2011caltech}, and CIFAR-FS\cite{bertinetto2018meta}.
All datasets follow a standard division and all images are resized to predefined resslutions following standard settings~\cite{zhang2020deepemd,rizve2021exploring}.
\textsl{Note that we do not introduce any external data.}

\noindent\textbf{{\em mini}ImageNet} was derived from ImageNet~\cite{russakovsky2015imagenet}, and has 100 diverse classes, including animal species and object categories, with 600 images per class.We use $64$, $16$, $20$ classes for training, validation and testing.

\noindent\textbf{{\em tiered}ImageNet} is another few-shot classification dataset built upon ImageNet~\cite{russakovsky2015imagenet}, containing 608 classes sampled from the hierarchical category structure. There are 351, 97 and 160 classes for training, validation and testing, respectively.

\noindent\textbf{CUB} was first proposed for fine-grained classification, and has 200 different bird species, among which 100 classes are used for training, 50 for validation and 50 for testing.

\noindent\textbf{CIFAR-FS}, a subset of CIFAR-100~\cite{krizhevsky2009learning}, was constructed by randomly splitting the 100 classes in the CIFAR-100 dataset into 64, 16 and 20 classes respectively for training, validation and testing. {All images are resized to predefined resoultions following standard settings~\cite{zhang2020deepemd,rizve2021exploring}}.

We evaluate the performance under standard 5-way-1-shot and 5-way-5-shot settings.
~\cite{kim2019edge,finn2017model}. 
\textsl{Note that we do not introduce any external data.}



\noindent{\bf{Implementation Details.}}
For the feature extractor, we consider three state-of-the-art backbones, S2M2~\cite{mangla2020charting}, IE~\cite{rizve2021exploring} and FEAT~\cite{ye2020few}. 
The number of neurons in different layers of RankMLP is $[6400,1024,512,256,1]$. There are only 6.8 million learnable parameters. We set the weight decay to $10^{-6}$ and the momenta to 0.9. The learning rate is fixed at 0.0005 for both networks. 
Note that during the meta test stage,  RankDNN does not need to finetune on 1-shot, but on 5-shot, RankMLP needs to be finetuned with the support set to get good performance, where we sample 100 triplets randomly in each mini-batch, and the parameters of RankMLP is updated in 100 iterations. The learning rate is set to 0.01. RankDNN is optimized with SGD.

\begin{table}[htpb]
\caption{Comparison of 5-way few-shot accuracies on CUB.}
	\centering
	\label{tab:cub}
	\scalebox{0.95}{
\begin{tabular}{lccc}
\bottomrule
\multirow{1}{*}{\bf {Method}} &\bf {Backbone} &\multicolumn{2}{c}{\bf CUB} \\
  \cmidrule(r){3-4} 
& &1-shot &5-shot\\
\hline
IE$_{\;\rm{\,CVPR21}}$
&ResNet-12&80.92 &90.13 \\
RENet$_{\;\rm{\,ICCV21}}$
&ResNet-12 &79.49 &91.11\\
HGNN$_{\;\rm{\,ICCV21}}$
&ResNet-12&78.58 &90.02\\
CCG+HAN$_{\;\rm{\,ICCV21}}$
&ResNet-12&74.66 &88.37\\
APP2S$_{\;\rm{\,AAAI22}}$
&ResNet-12&77.64 &90.43\\
{\bf RankDNN(ours)} &ResNet-12&{\bf 82.93} &{\bf 91.47}\\
\hline
S2M2$_{\;\rm{\,WACV20}}$
&WRN &80.68 &90.85\\
DC$_{\;\rm{\,ICLR21}}$
&WRN&79.56 &90.67\\
{\bf RankDNN(ours)} &WRN&{\bf 81.78}
&{\bf 91.12}\\
\bottomrule

\end{tabular}}
\label{tab:cub}
\end{table}

\noindent{\bf Ranking Voting.} For an $N$-way-$K$-shot task, there are a total of $NK$ support images. After feature extraction, we average the $K$ features to one per class, so we get $N$ support features whether on the 1-shot or n-shot setting. Each query feature is treated as an anchor, and every two support out of the $N$ average features can be used to form a triplet with the query.
Thus, $N \times (N-1)$ valid triplets can be constructed for each query. If a triplet is predicted positive by our RankDNN, the first support image in the triplet is given one positive point. We define the ranking score of a support sample to be the total number of positive points received by the support sample. The class label of the query image is the same as the class label of the support sample with the highest ranking score.


\begin{table}[htpb]
\caption{Comparison of 5-way few-shot on CIFAR-FS.}
	\centering
	\label{tab:cifar}
	\scalebox{0.95}{
\begin{tabular}{lccc}
\bottomrule
\multirow{1}{*}{\bf {Method}} &\bf {Backbone} &\multicolumn{2}{c}{\bf CIFAR-FS} \\
  \cmidrule(r){3-4} 
& &1-shot &5-shot\\
\hline
IE$_{\;\rm{\,CVPR21}}$
&ResNet-12&77.87 &89.74 \\
PAL$_{\;\rm{\,ICCV21}}$
&ResNet-12&77.10 &88.00 \\
TPMN$_{\;\rm{\,ICCV21}}$
&ResNet-12&75.50 &87.20 \\
CCG+HAN$_{\;\rm{\,ICCV21}}$
&ResNet-12&73.00 &85.80 \\
APP2S$_{\;\rm{\,AAAI22}}$
&ResNet-12&73.12 &85.69 \\
LH$_{\;\rm{\,AAAI22}}$
&ResNet-12&78.00 &90.50 \\
{\bf RankDNN(ours)} &ResNet-12&{\bf 78.93} &{\bf 90.64}\\
\bottomrule
\end{tabular}}
	\label{tab:cifar}
\end{table}

\begin{table*}[thpb]
\caption{Comparison of fusing RankDNN with other methods  on {\em mini}ImageNet and {\em tiered}ImageNet.}
	\centering
	\label{tab:fusion mini/tiered}
	\scalebox{0.85}{
\begin{tabular}{lccccc}
\bottomrule
\multirow{1}{*}{\bf {Method}} &\bf {Backbone} &\multicolumn{2}{c}{\bf {{\em mini}ImagNet}} &\multicolumn{2}{c}{\bf {{\em tiered}Imagenet}} \\
  \cmidrule(r){3-4}  \cmidrule(r){5-6} 
& &1-shot &5-shot &1-shot &5-shot\\
\hline
ProtoNet~\cite{snell2017prototypical}
&ResNet-12 &61.20 &77.55 &68.01 &83.91\\
{\bf ProtoNet+RankDNN} &ResNet-12&64.75$_{{\color{red}+3.55}}$ &79.01$_{{\color{red}+1.46}}$ &70.12$_{{\color{red}+2.01}}$ &85.00$_{{\color{red}+1.09}}$\\
\hline
E$^{3}$BM~\cite{liu2020ensemble}
&ResNet-12 &63.80 &80.10 &71.20 &85.30\\
{\bf E$^{3}$BM+RankDNN} &ResNet-12 &65.00$_{{\color{red}+1.20}}$ &80.45$_{{\color{red}+0.35}}$ &71.72$_{{\color{red}+0.52}}$ &86.22$_{{\color{red}+0.92}}$\\
\hline
DeepEMD~\cite{zhang2020deepemd}
&ResNet-12 &66.50 &82.41 &72.65 &86:03\\
{\bf DeepEMD+RankDNN} &ResNet-12 &67.01$_{{\color{red}+0.51}}$ &84.32$_{{\color{red}+1.91}}$ &72.80$_{{\color{red}+0.15}}$ &86.10$_{{\color{red}+0.07}}$\\
\hline
Distill~\cite{tian2020rethinking}
&ResNet-12 &64.82 &82.14 &71.52 &86.03\\
{\bf Distill+RankDNN} &ResNet-12 &66.34$_{{\color{red}+1.52}}$ &83.98$_{{\color{red}+1.48}}$ &72.20$_{{\color{red}+0.68}}$ &86.45$_{{\color{red}+0.42}}$\\
\hline
FRN~\cite{wertheimer2021few}
&ResNet-12 &66.25 &82.50 &72.06 &86.37\\
{\bf FRN+RankDNN} &ResNet-12 &66.58$_{{\color{red}+0.33}}$ &82.98$_{{\color{red}+0.48}}$ &72.21$_{{\color{red}+0.15}}$ &86.62$_{{\color{red}+0.25}}$\\
\hline
RENet~\cite{kang2021relational}
&ResNet-12 &67:60 &82:58 &71:61 &85:28\\
{\bf RENet+RankDNN} &ResNet-12 &68.01$_{{\color{red}+0.41}}$ &84.00$_{{\color{red}+1.42}}$ &71.85$_{{\color{red}+0.24}}$ &85.98$_{{\color{red}+0.70}}$\\
\hline
DC~\cite{yang2021free}
&WRN &68.57&82.88 &78.19 &89.90\\{\bf DC+RankDNN}&WRN &69.54$_{{\color{red}+0.97}}$ &83.66$_{{\color{red}+0.78}}$ &78.92$_{{\color{red}+0.73}}$ &90.20$_{{\color{red}+0.30}}$\\

\bottomrule

\end{tabular}}

	\label{tab:fusion mini/tiered}
\end{table*}

\subsection{Comparisons with the State-of-the-arts}
As shown in Table~\ref{tab:mini_and_tiered}, Table~\ref{tab:cub} and Table~\ref{tab:cifar}, RankDNN outperforms its baselines obviously on all benchmarks. It demonstrates that the relevance ranking is helpful to distinguish similar images. 
Also, RankDNN surpasses all previous state-of-the-art algorithms.
On the {\em mini}ImageNet,when compared with the previous state-of-the-art methods(IE on ResNet-12, FEAT on ResNet-18 and S2M2 on WRN), RankDNN achieves $1.44$ \%, $0.88$\%, $0.88$\%,$5.47$\%,$0.49$\%,$1.74$\% and $1.61$\% improvements in 5-way-1-shot and 5-way-5-shot accuracies respectively. We attribute this to the strong generalization ability of RankDNN, which can be used to enhance various few-shot learning algorithms.
On {\em tiered}ImageNet, RankDNN shows obvious performance improvements over the S2M2, FEAT and IE baselines under both 5-way-1-shot and 5-way-5-shot settings. RankDNN with the ResNet-12 backbone surpasses previously best performing TAS by 1.06\% and 1.86\% under both 5-way few-shot settings respectively. RankDNN with the ResNet-18 backbone achieves new state-of-the-art performance ($66.97$\% and $81.07$\%) under both settings. 
RankDNN with the WRN backbone also surpasses previously best performing S2M2.
This proves the effectiveness and robustness of our RankDNN.

On the CUB dataset, RankDNN with both S2M2 and IE backbones achieves impressive performance under both 5-way few-shot settings (Table~\ref{tab:cub}). 
As CUB is a dataset for fine-grained classification, such performance indicates that RankDNN has a very strong ability to distinguish subtle differences in visual semantics. 



On the CIFAR-FS, where images have a very low resolution, RankDNN with the IE backbone surpasses its baseline, which is also the recent best, by 1.14\% and 0.90\% under 5-way-1-shot and 5-way-5-shot settings, respectively (Table ~\ref{tab:cifar}). These results indicate RankDNN are complementary to state-of-the-art methods, and can generalize well on previously unseen visual concepts.

\subsection{Comparisons of the State-of-the-arts and their fusion with RankDNN}
We reproduce the state-of-the-arts methods, including ProtoNet~\cite{snell2017prototypical},
E$^{3}$BM~\cite{liu2020ensemble}, DeepEMD~\cite{zhang2020deepemd}, Distill~\cite{tian2020rethinking}, FRN~\cite{wertheimer2021few}, RENet~\cite{kang2021relational}and DC~\cite{yang2021free}),
and fusion them with RankDNN.
As shown in Table~\ref{tab:fusion mini/tiered},
fusion with RankDNN improves the performance of all original methods, both on 1-shot and 5-shot.
In particular, for ProtoNet, RankDNN improves $3.55$\% and $1.46$\% on {\em mini}ImageNet and 
$2.01$\% and $1.09$\% on {\em tiered}ImageNet.
These results indicate that RankDNN can be flexibly adapted to different baselines, and various methods can benefit from RankDNN.

\begin{figure}[h]
	\centering
	\includegraphics[width=1.0\linewidth]{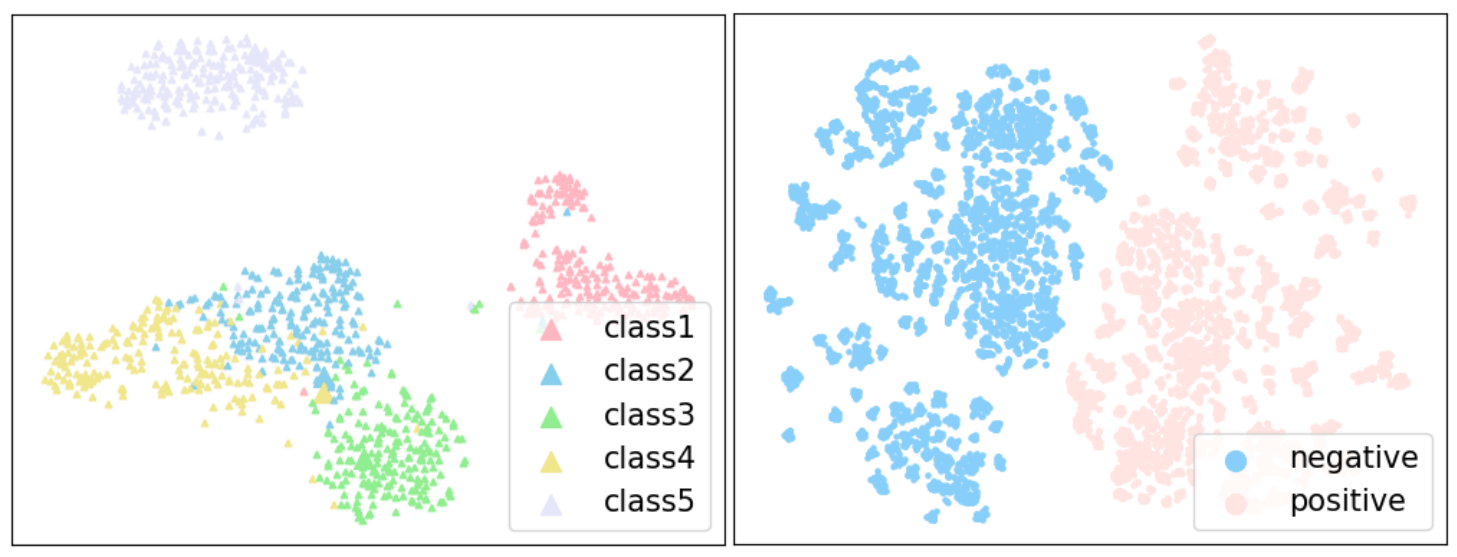}
	\caption{For a 5-way-100-shot task, we visualize the feature descriptors generated by the feature backbone and RankMLP with t-SNE.}
	\label{Fig:tsne} 
\end{figure}

\subsection{Discriminative Ability of Feature Descriptors in RankDNN} 
We claim that features of the rank-triples formed by the Kronecker are more discriminative and stable than the original deep features. To validate, for a 5-way-100-shot task, we visualize features extracted by the backbone and the second last layer of RankMLP with t-SNE in Fig~\ref{Fig:tsne}. It can be found that backbone features will result some ambiguity around the classification boundaries. However, the features generated by RankMLP have a stronger discriminative ability in distinguishing positive and negative ranked samples.

\begin{table}[htpb]
	\caption{Ablation study on {\em mini}ImageNet and CUB using two baselines and under the 5-way-1-shot setting. }
\centering
\label{tab:abl}
\scalebox{0.95}{
\begin{tabular}{lcc}
\bottomrule
\multirow{1}{*}{\bf Model}  &\multicolumn{1}{c}{\bf {\em mini}ImageNet} &\multicolumn{1}{c}{\bf CUB} \cr
\hline
IE~\cite{rizve2021exploring}&67.15 &80.92  \\
After PCA &65.32 &80.11 \\
Feature Differentiation &60.57$\downarrow$ &77.33$\downarrow$\\
Triple Concat &20  &20 \\
Pairwise Concat  &42.50$\downarrow$ &44.65$\downarrow$ \\
Hadamard  &65.00$\downarrow$ &81.64$\uparrow$ \\
Kronecker &\bf 68.72$\uparrow$ &\bf 82.93$\uparrow$ \\
Polynomial &21.00$\downarrow$  &22.00$\downarrow$\\
			\hline
S2M2~\cite{mangla2020charting}&64.50 &80.68\\
After PCA &65.32 &80.11 \\
Kronecker+RankSVM &60.1$\downarrow$ &78.24$\downarrow$ \\
Feature Differentiation  &61.24$\downarrow$ &77.83$\downarrow$\\
Hadamard   &64.22 &79.93$\uparrow$ \\
Kronecker  &66.67$\uparrow$ &\bf 81.78$\uparrow$ \\
The Combined Product &\bf 66.88$\uparrow$  & 81.50$\uparrow$ \\
\bottomrule
	\end{tabular}}	

\label{tab:abl}
\end{table}

\subsection{Ablation Study}
\noindent{\bf{Dimension Reduction Module.}} {In Table~\ref{tab:abl}, the 5-way-1-shot/5-way-5-shot classification results of baselines and baselines after the dimension reduction module are obtained from cosine classifiers and linear regression classifiers respectively. It indicates that PCA method will slightly decrease the performance of baselines. Meanwhile, RankDNN can improve the classification accuracies efficiently.}

\noindent{\bf{Choices in RankMLP.}} 
{As in Table~\ref{tab:abl},first, we verify the effectiveness of RankMLP by replacing RankMLP with RankSVM~\cite{joachims2002optimizing} on S2M2, and the performance on miniImageNet and CUB drops by $4.49\%$ and $2.44\%$ respectively. Second, we explore different triplet encoding schemes by replacing the vector-Kronecker product with feature disparity (Eq.~\ref{eq:rank_description_man}), feature concatenation of triplets, feature disparity of pairwise concatenation, Hadamard product (Eq.~\ref{eq:rank_description_Had}), 
	and the combination of Kronecker and Hadamard products (Eq.~\ref{eq:rank_description_com}).
We can see that the Kronecker product has the best generalization performance and is applicable to all backbones and datasets.
The feature disparity scheme does not well preserve the original feature and sign information and therefore performs poorly.
Simply concatenating three feature vectors of a triplet will lead to gradient explosion, and the feature disparity of pairwise concatenation works in low efficiency. 
Polynomial product fails to describe the similarity information within the triplets, resulting in the subsequent network not being able to learn ranking information.
Hadamard product is essentially dot product for vectors and has similar performance as the baseline. However, the combination of Kronecker and Hadamard products can surpass Kronecker product alone in some cases.

\begin{equation}
	\psi \left (  \left \langle x_q,x_{i},x_{j} \right \rangle; \theta \right ) = \left| {f_q} - {f_i}\right| -  \left| {f_q} - {f_j}\right|.
	\label{eq:rank_description_man}
\end{equation}
\begin{equation}
	\psi \left (  \left \langle x_q,x_{i},x_{j} \right \rangle; \theta \right ) = {f_q} \odot  {f_i} - {f_q}  \odot  {f_j}.
	\label{eq:rank_description_Had}
\end{equation}
\begin{equation}
	\psi \left (  \left \langle x_q,x_{i},x_{j} \right \rangle; \theta \right ) = ({f_q} \otimes {f_i};{f_q} \odot {f_i}) - ({f_q}  \otimes  {f_j};{f_q} \odot {f_j}).
	\label{eq:rank_description_com}
\end{equation}

\section{Conclusions}
\label{Conclusions}
In this paper, we have introduced RankDNN, a novel pipeline for few-shot learning. It can convert a multiclass classification task into a binary ranking relation classification problem.
Accurate binary ranking relation classification is made possible by an informative encoding scheme of image triplets and later uses the  simple fully connected network to predict whether there are in a query-relevant-irrelevant order or not.
Experiments demonstrate the proposed method achieves new state-of-the-art performance on four benchmarks. Meanwhile, experimental results for different backbones and cross-domain settings demonstrate RankDNN is data efficient and domain agnostic.

\section{Acknowledgments}

\bigskip
\noindent This work was supported by National Key R\&D Program of China (2020AAA0108301), National Natural Science Foundation of China (No.62072112 and No.62106051), Scientific and Technological innovation action plan of  Shanghai Science and Technology Committee (No.22511102202), Fudan University Double First-class Construction Fund (No.XM03211178), and the Shanghai Pujiang Program (No.21PJ1400600).

\section{Supplementary Material}

\subsection{Additional Experiments}
\subsubsection{Applicability}
We tested the applicability of our method to a variety of backbone networks, including Conv-4~\cite{vinyals2016matching}, Conv-6\cite{chen2019closer}, ResNet-12~\cite{rizve2021exploring}, ResNet-18~\cite{ye2020few}, and WRN-28~\cite{mangla2020charting}. These backbones serve as the feature extractor in our pipeline. 

\begin{table}[htpb]
\caption{5-way few-shot classification accuracies of RankDNN on {\em mini}ImageNet using different backbones.}
	\centering
	\label{tab:backbones}
	\scalebox{0.73}{
\begin{tabular}{lcccc}
\bottomrule
\multirow{1}{*}{\bf {Backbone}}  &\multicolumn{2}{c}{\bf \textit{mini}Imagenet} &\multicolumn{2}{c}{\bf with RankDNN}\cr
&1-shot &5-shot &1-shot &5-shot\\
\hline
Conv-4~\cite{vinyals2016matching} &43.56 &53.11 &54.20$_{{\color{red}+10.64}}$&66.00$_{{\color{red}+12.89}}$ \\
Conv-6~\cite{chen2019closer} &46.00 &61.50 &56.16$_{{\color{red}+10.16}}$&66.77$_{{\color{red}+5.27}}$ \\
ResNet-12~\cite{rizve2021exploring} &67.28 &84.78 &68.72$_{{\color{red}+1.44}}$&85.66$_{{\color{red}+0.88}}$ \\
ResNet-18~\cite{ye2020few} &55.15 &66.78 &62.52$_{{\color{red}+7.37}}$&77.33$_{{\color{red}+10.55}}$ \\
WRN~\cite{mangla2020charting} &64.93 &83.18 &66.67$_{{\color{red}+1.74}}$&84.79$_{{\color{red}+1.61}}$ \\
\bottomrule

\end{tabular}}
\label{tab:backbones}
\end{table}

As shown in Table~\ref{tab:backbones}, in comparison to the bare backbones, our RankDNN brings over 5\% accuracy improvement on Conv-4, Conv-6 and ResNet-18 on both 1-shot and 5-shot.
Note that the comparison baseline used in conv-4 is MatchingNet~\cite{vinyals2016matching}, where the performance of RankDNN improves 10.64\% on 1-shot and 12.80\% on 5-shot, where it can be proved that RankDNN has superior performance compared with the matching methods in MatchingNet.
It also brings over 1\% improvement on WRN-28 and ResNet-12. 
RankDNN can improve the performance of various backbones, especially on 1-shot setting. These results demonstrate the strong robustness of RankDNN.
\begin{table}[htpb]
\caption{Ablations on learnable parameters of RankDNN.}
\Large
 \centering
 \scalebox{0.51}{
  \begin{tabular}{lcccc}
\bottomrule
 \multirow{1}{*}{\bf Model}  
 &\multicolumn{1}{c}{\bf RankMLP Configurations} &\multicolumn{1}{c}{\bf Parameters(MB)}
 &\multicolumn{2}{c}{\bf \textit{mini}Imagenet}\cr
 & & &1-shot &5-shot\\
    \hline
     Baseline IE &-- &-- &$67.28$ &$84.78$\\
    RankDNN-Lite &$(4096,1024,512,256,1)$ &$4.50$ &$68.15$ &$85.01$\\
    RankDNN &$(6400,1024,512,256,1)$ &$6.80$ &$68.72$ &$85.66$\\
    RankDNN-Large1  &$(6400,2048,1024,512,1)$ &$14.50$ &$68.83$ &{\color{red}$85.88$}\\
    RankDNN-Large2 &$(16384,4096,2048,1024,512,1)$ &$74.50$ &$71.25$ &$85.06$\\
    RankDNN-Large3  &$(16384,8000,4000,2000,1000,1)$ &$173.09$ &{\color{red}$72.07$} &$85.11$\\
\bottomrule
 \end{tabular}}
 \label{tab:abl_param}
\end{table}
\subsubsection{Ablation Study on Learnable Parameters and Layers of RankDNN}
After feature extraction, we downscaled the original features by different PCA to obtain three input dimensions of RankDNN, including $16,384$, $6,400$, and $4,096$.
We designed the corresponding MLPs respectively to explore the relationship between the number of parameters and performance in RankDNN.

\begin{table*}[thpb]
\caption{Comparison of the number of model parameters.}
\Large
 \centering
 \label{tab:param}
 \scalebox{0.77}{
  \begin{tabular}{lccccccc}
\bottomrule
 \multirow{1}{*}{\bf Model}  
 &\multicolumn{1}{c}{\bf Backbone} 
 &\multicolumn{1}{c}{\bf Total Param} 
 &\multicolumn{1}{c}{\bf Backbone Param} 
 &\multicolumn{1}{c}{\bf Head Param}
  &\multicolumn{1}{c}{\bf GFlops}
 &\multicolumn{2}{c}{\bf \textit{mini}Imagenet}\cr
   \cmidrule(r){7-8}  
 & &(MB) &(MB) &(MB) & &1-shot &5-shot \\
    \hline
    \bf RankDNN(ours) &ResNet-12 &19.38 &12.47 &6.87 &182.61&68.72 &85.66\\
     XtarNet~\cite{yoon2020xtarnet}  &ResNet-12 &19.87 &12.47 &7.40 &175.92&49.71 &60.13\\
     MTL~\cite{sun2019meta}  &ResNet-12 &16.67 &12.47 &4.20 &175.92&$61.20$ &$71.50$\\
     wDAE~\cite{gidaris2019generating}  &WRN &47.52 &36.51 &11.01 &1808.08&$62.96 $ &$78.85$\\
     FEAT~\cite{ye2020few}  &ResNet-18 &34.8 &33.2 &1.6 &175.92&$66.78$ &$82.05$\\
\bottomrule
 \end{tabular}}
  \label{tab:param}
\end{table*}

\begin{table}[htpb]
 \caption{Ablation study on different choices of the layer numbers in RankMLP. Results of 5-way-1-shot classification are reported.}
 \centering
\label{tab:layers}
 \large
 \scalebox{0.80}{
  \begin{tabular}{llcc}
     \bottomrule
     \specialrule{0em}{2pt}{2pt}
 \multirow{1}{*}{\bf MLPs} &{\bf Configurations} &{\bf {\em mini}} &{\bf{\em tiered}}\\
 \specialrule{0em}{2pt}{2pt}
   \hline
   \specialrule{0em}{2pt}{2pt}
   Baseline &ResNet-12 &67.28 &72.21\\
   One-layer &$(6400,1)$&65.97 &70.45\\
   Three-layers &$(6400,512,256,1)$&68.44 &73.01\\
   Four-layers&$(6400,1024,512,256,1)$&68.72 &73.87 \\
   Five-layers &$(6400,2048,1024,512,256,1)$&68.78 &73.92 \\
   \specialrule{0em}{2pt}{2pt}
     \bottomrule
 \end{tabular}}
 \label{tab:layers}
\end{table}

As in Table~\ref{tab:abl_param}, 
RankDNN with $173.09M$ parameters get the best result $72.07\%$ on 1-shot. It proves that more learnable parameters in RankMLP lead to better 1-shot classification results.  
But for 5-shot classification tasks, RankDNN-Large1 performs better.
Besides, we found that RankDNN with a large number of parameters cannot continue to improve its performance by fine-tuning on 5-shot tasks.
It indicates that RankMLP needs a lot of parameters to learn ranking relationships. But too many parameters still lead to over-fitting during meta-test.
The RankDNN with $6.8M$ parameters performs well in both 1-shot and 5-shot tasks and does not consume large amounts of memory.

Also, we tested the choices of different layer numbers in RankMLP. As shown in Table~\ref{tab:layers}, 
RankMLP using only one-layer cannot learn the ranking relationship well, resulting in lower performance than the baseline.
The four-layers MLPs achieve better performances than their three-layers counterparts, especially on {\em tiered}ImageNet. 
Compared with the four-layers, the five-layers one has no significant improvement but the number of parameters will be increased by about $6M$ parameters, so the four-layer RankDNN is chosen in the paper.

\subsubsection{Ablation Study on Different Testing Methods}
During the meta test, we did not fine-tune RankDNN on the 5-way 1-shot setting. However, due to the unique triplets construction, we can construct positive and negative samples of $2,000$ respectively for fine-tuning RankDNN online on the 5-way 5-shot setting.
Therefore, we compared the results before and after fine-tuning, 
 as in Table ~\ref{tab:testing}.
Experiments show that without finetuning, results of RankDNN are even lower than its baseline and after fine-tuning, RankDNN can improve about $2\%$ on 5-way 5-shot on {\em mini}ImageNet, which demonstrates the validity of the ranking triplets constructed from the Kronecker product and the rapid learning capability of RankDNN.
Meanwhile, in order to reduce the test time of ranking voting on K-shot tasks, we average the support features from the same class to one for ranking. And the experiments show that the averaging operation not only reduces the testing time by $>1$s but also improves the performance of RankDNN.

\begin{table}[htpb]
 \caption{Ablation study on different testing methods of 5-shot task on {\em mini}ImageNet.}
 \centering
\label{tab:testing}
 \large
 \scalebox{0.75}{
  \begin{tabular}{lcccc}
     \bottomrule
     \specialrule{0em}{2pt}{2pt}
 \multirow{1}{*}{\bf Method} &{\bf Fine-tune} &{\bf Average} &{\bf Test time(s)} &{\bf {\em mini}ImageNet} \\
 \specialrule{0em}{2pt}{2pt}
   \hline
   \specialrule{0em}{2pt}{2pt}
   Baseline &$\times$ &$\times$ &0.065 &84.78\\
  RankDNN &$\times$ &$\times$ &1.52 &83.20\\
  RankDNN &\checkmark &$\times$ &7.58 &85.42\\
  RankDNN &$\times$ &\checkmark &0.41 &83.87\\
  RankDNN &\checkmark &\checkmark &6.46 &85.66\\
   \specialrule{0em}{2pt}{2pt}
     \bottomrule
 \end{tabular}}
 \label{tab:testing}
\end{table}

\subsubsection{Comparison of Learnable Parameters, FLOPs and Accuracies}
RankDNN contains a feature backbone and a classification head that increasing  6.87M parameters compared to the feature extraction methods.
According to Table~\ref{tab:param},
compared with other state-of-the-art post-processing methods ,the increased number of parameters and GFlops of RankDNN are in a reasonable range, but the performance  of RankDNN is more outstanding.

\begin{table}[htpb]
 \caption{Performance comparison in the cross-domain setting: {\em mini}ImageNet$\rightarrow$CUB.}
\normalsize
 \centering
 \label{tab:freq}
 \scalebox{1.0}{
  \begin{tabular}{lcc}
\bottomrule
 \multirow{1}{*}{\bf Model}  &\multicolumn{2}{c}{\bf {\em mini}ImageNet $\rightarrow$ \bf CUB} \cr
  &\bf 1-shot &\bf 5-shot \\
   \hline
   SimpleShot $_{\;\rm{\,PMLR20}}$  &$48.56_{\pm 1.90}$ &$65.63_{\pm 1.00}$ \\
    GNN+FT$_{\;\rm{\,ICLR20}}$  &$47.47_{\pm 0.90}$ &$66.98_{\pm 0.60}$ \\
    Afrasiyabi $_{\;\rm{\,ECCV20}}$  &$46.85_{\pm 0.38}$ &$70.37_{\pm 0.35}$ \\
    MeTAL$_{\;\rm{\,ICCV20}}$  &-- &$70.22_{\pm 0.33}$ \\
    FRN$_{\;\rm{\,CVPR20}}$ &$53.39_{\pm 0.47}$ &$75.16_{\pm 0.33}$ \\
    IE$_{\;\rm{\,CVPR21}}$ &$53.56_{\pm 0.78}$ &$75.67_{\pm 0.86}$ \\
   \bf RankDNN(ours) &${
   \bf 60.64_{\pm 0.22}}$ & ${\bf 85.92_{\pm 0.24}}$  \\
\bottomrule
 \end{tabular}}
 \label{tabel2}
\end{table}

\subsubsection{Cross-Domain Few-Shot Classification}
We conducted experiments in the cross-domain challenge and compared them with some recent works~\cite{wang2019simpleshot,chauhan2020few} et al, where the model is trained on the {\em mini}ImageNet and evaluated on the CUB. In this setting, as shown in Table.\ref{tabel2}, RankDNN improves  7.08$\%$ on 1-shot classification tasks and 10.25\% on 5-shot classification tasks over the baseline and significantly exceeds the performance of other methods. These results demonstrate ordering relations within triplets are widely present in all categories and RankDNN has superior migratability and domain agnostic.

\begin{table}[htpb]
 \caption{Performance comparison in other cross-domain settings.}
\Large
 \centering
 \label{tab:cross2}
 \scalebox{0.63}{
  \begin{tabular}{lccc}
\bottomrule
 \multirow{1}{*}{\bf Model} &\multirow{1}{*}{\bf CIFAR $\rightarrow$ \bf CUB}  &\multirow{1}{*}{\bf CIFAR $\rightarrow$ \bf {\em mini}} &\multirow{1}{*}{\bf {\em mini} $\rightarrow$ \bf CIFAR}\cr
&\bf 1-shot/5-shot  &\bf 1-shot/5-shot &\bf 1-shot/5-shot \\
\hline\rule{0pt}{15pt}
Baseline IE  &$48.91/58.25$ &$55.45/70.45$ &$63.77/78.63$\\
\hline\rule{0pt}{15pt}
\bf RankDNN(ours)  &$54.01/65.00$ &$55.33/74.29$ &$64.44/80.97$\\
\bottomrule
 \end{tabular}}
 \label{tab:cross2}
\end{table}

 We also give results of the other three settings, including CIFAR$\rightarrow$ CUB, CIFAR$\rightarrow$ {\em mini}ImageNet and {\em mini}Imagenet $\rightarrow$ CIFAR.
Compared with the baseline in Table.\ref{tab:cross2}, most of the results are improved,especially on the 5-shot setting.
This is because in a 5-way 5-shot meta test, RankDNN can adapt the ranking relationships of new categories through fast iterations with a lot of rank samples constructed online.
These results demonstrate that RankDNN can learn a unified relevance ranking classifier across different domains and have fast online learning capability for new classes.

\subsection{Supplementary References}
The references not cited in Table 1, 2 and 3 in the main paper are as follows: FRN~\cite{wertheimer2021few}, DMF~\cite{xu2021learning}, IEPT~\cite{zhang2020iept}, DMN4~\cite{liu2022dmn4}, HGNN~\cite{yu2022hybrid}, APP2S~\cite{ma2022adaptive}, UNICORN~\cite{ye2021train}, TAS~\cite{le2021task}, $\bigtriangleup-$encode~\cite{schwartz2018delta}, SS~\cite{su2020does}, Hyperbolic~\cite{khrulkov2020hyperbolic}, PSST~\cite{chen2021pareto}, CCG+HAN~\cite{gao2021curvature}, PAL~\cite{ma2021partner}, TPMN~\cite{wu2021task}, LH~\cite{jian2022label}.

\bibliography{aaai23}

\end{document}